\documentclass[journal, caption=false]{IEEEtai}

\usepackage[colorlinks,urlcolor=blue,linkcolor=blue,citecolor=blue]{hyperref}

\usepackage{graphicx} 
\usepackage[caption=false,font=footnotesize]{subfig}
\usepackage{colortbl} 

\usepackage{tabularx}

\usepackage[usenames,dvipsnames,svgnames,table, x11names]{xcolor}
\usepackage{multirow}

\usepackage{comment}

\usepackage{tikz}
\usetikzlibrary{arrows.meta, shapes.geometric, positioning}

\usepackage{amsmath}
\usepackage{amssymb}
\usepackage{mathtools}
\usepackage{amsthm}

\usepackage[capitalize,noabbrev]{cleveref}

\theoremstyle{plain}

\theoremstyle{definition}

\theoremstyle{remark}

\usepackage{fontawesome5} 

\newcommand{\senioricon}[1][1.2em]{%
    \includegraphics[height=#1]{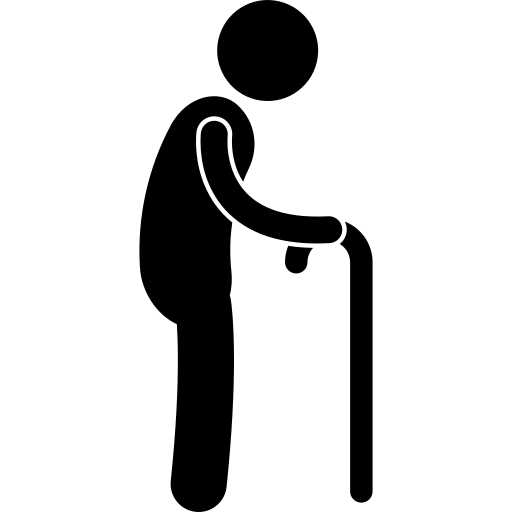}%
}

\newcommand{\ceilingmount}[1][1.2em]{%
    \includegraphics[height=#1]{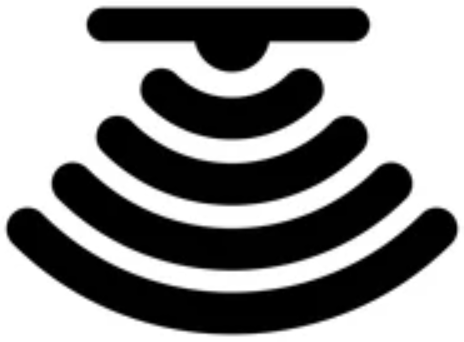}%
}

\newcommand{\fall}[1][2em]{%
    \includegraphics[trim=0.1cm 0.1cm 0.1cm 0.2cm, clip, height=#1]{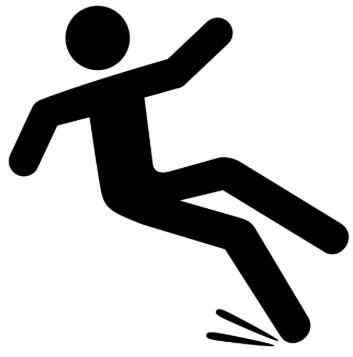}%
}

\usepackage[textsize=tiny]{todonotes}

\begin{document}

\title{Thermal Imaging-based Real-time Fall Detection using Motion Flow and Attention-enhanced Convolutional Recurrent Architecture}

\author{
Christopher Silver$^{1}$ and Thangarajah Akilan$^{2}$ \\
$^{1}$Department of Electrical and Computer Engineering, Lakehead University, Thunder Bay, Ontario, Canada \\
$^{2}$Department of Software Engineering, Lakehead University, Thunder Bay, Ontario, Canada \\
\texttt{crsilver@lakeheadu.ca, takilan@lakeheadu.ca}
}


\maketitle

\begin{abstract}
Falls among seniors are a major public health issue. Existing solutions using wearable sensors, ambient sensors, and RGB-based vision systems face challenges in reliability, user compliance, and practicality. Studies indicate that stakeholders, such as older adults and eldercare facilities, prefer non-wearable, passive, privacy-preserving, and real-time fall detection systems that require no user interaction. This study proposes an advanced thermal fall detection method using a Bidirectional Convolutional Long Short-Term Memory (BiConvLSTM) model, enhanced with spatial, temporal, feature, self, and general attention mechanisms. Through systematic experimentation across hundreds of model variations exploring the integration of attention mechanisms, recurrent modules, and motion flow, we identified top-performing architectures. Among them, BiConvLSTM achieved state-of-the-art performance with a ROC-AUC of $99.7\%$ on the TSF dataset and demonstrated robust results on TF-66, a newly emerged, diverse, and privacy-preserving benchmark. These results highlight the generalizability and practicality of the proposed model, setting new standards for thermal fall detection and paving the way toward deployable, high-performance solutions.

\end{abstract}

\begin{IEEEImpStatement}
Falls remain a critical concern for older adults, necessitating real-time detection systems that are both accurate and privacy-preserving. This study introduces a novel thermal-based approach, designed to protect user dignity while ensuring robust fall detection in diverse environments. Through extensive evaluation of architectural variants, the work identifies practical trade-offs between accuracy, efficiency, and latency, confirming that real-time feasibility can be achieved without sacrificing reliability.
A key contribution is the emphasis on dataset diversity, demonstrated through strong generalization on the newly introduced TF-66 benchmark. Beyond technical performance, the system is envisioned as an AI-assisted tool to support caregivers and help mitigate the growing shortage of personal support workers, thereby extending the capacity of eldercare facilities. Ultimately, this work advances the field toward deployable fall detection systems that enhance safety, autonomy, and acceptance among at-risk populations.
\end{IEEEImpStatement}


\section{Introduction}

\begin{table*}[ht]
\centering
\setlength{\tabcolsep}{3.0pt} 
\caption{A summary of the key fall detection works on the TSF dataset. The results identify the best ROC-AUC reported}
\label{tab:lit-unsupervised}
\vspace{-0.6cm}
\begin{flushleft} 
\end{flushleft}
\begin{small}
\begin{tabular}{>{\centering\arraybackslash}m{0.5cm} 
                >{\raggedright\arraybackslash}m{8.3cm} 
                >{\raggedright\arraybackslash}m{7.6cm} 
                >{\raggedright\arraybackslash}m{0.8cm}
                }
\hline

    {Ref.}
    & \centering \textbf{Methodology}
    & \centering \textbf{Limitations} 
    & {Results} \\ 
    \hline\hline
  
 \cite{nogas2018fall}& A ConvLSTM-AE on visual-enhanced thermal ADL images, identifying high reconstruction scores of data as anomalies indicating falls. & The system had mediocre results, not being reliable enough for real-world implementation.    & 83.0\%
    \\ \hline
\cite{nogas2020deepfall}& A 3D Convolutional Autoencoder. & Small input window size misses vital temporal contexts.   & 97.0\%
    \\ \hline
   \cite{elshwemy2020new} & Uses a Spatiotemporal Residual Autoencoder, using convolutional layers for spatial features, ConvLSTMs for temporal features, residual connections for efficiency. & Only use 8 frames per sample. Dataset is 12fps, meaning only 0.67 seconds are captured each classification, missing vital temporal contextual information.   & 97.0\% 
    \\ \hline
  \cite{khan2021spatio} & Employs a spatio-temporal adversarial framework consisting of a 3D convolutional autoencoder for reconstructing sequences of ADL and a 3D CNN as a discriminator. & System had worse results than previous solutions for TSF. The adversarial framework's reliance on reconstruction and discrimination adds computational overhead.   & 95.0\%
    \\ \hline
 
    \cite{mehta2021motion} & A dual-channel adversarial network processes thermal frames and motion flow. They also added region of interest extraction.  & Poor performance and reliance on accurate ROI extraction can fail with occlusion or tracking errors.    & 93.0\%
    \\ \hline
    \cite{silver2023novel} & Integration of a 3D CNN and an AE through a meta-model. Combining supervised and unsupervised models. & Sub-models were not optimized, causing mediocre results not reliable enough for real-world implementation.   & 83.0\%
    \\ \hline \hline
\end{tabular}
\end{small}
\end{table*}

\IEEEPARstart{W}{orld} Health Organization (WHO) defines a fall as ``inadvertently coming to rest on the ground or a lower level, excluding intentional changes in position''\footnote{\url{https://www.who.int/news-room/fact-sheets/detail/falls/}}. Falls are a leading cause of injury and mortality worldwide~\cite{yu2022data, chen2024fall}, particularly among seniors as the global population ages. Falls account for nearly half of all accidental injuries in seniors, often resulting in significant physical, psychological, and financial consequences~\cite{chaudhuri2017older, wang2023convolution}. With the number of individuals aged 65 and older projected to grow substantially between 2020 and 2050, the prevalence and impact of falls are expected to increase~\cite{alam2022vision, naser2022privacy}. Real-time, automatic FDS have become critical technologies, enhancing seniors' independence while reducing burdens on caregivers and healthcare systems~\cite{rafferty2019thermal}. Existing FDS solutions face significant challenges, including limited generalizability due to small, non-diverse datasets~\cite{riquelme2019ehomeseniors}, high false alarm rates, and difficulties in achieving real-time performance~\cite{newaz2023methods}. 
Video-based systems using RGB data raise privacy concerns, limiting their real-world applicability~\cite{pentyala2021privacy}.
A recent focus group study~\cite{silver2025thermal} further revealed that privacy preservation was the primary concern, cited by $88\%$ of senior participants in FDS.
This work addresses these limitations by systematically exploring and evaluating hundreds of model variations to develop an optimal thermal imaging-based fall detection solution using BiConvLSTM with attention mechanisms.
The proposed models achieve state-of-the-art results on the TSF~\cite{mehta2021motion} dataset and establish new benchmarks on TF-66~\cite{silver2025thermal}, a recently emerged ceiling-mounted thermal imaging dataset designed for human fall detection in elder care. 
Additionally, the models demonstrate real-time feasibility, bridging the gap between research and practical deployment.

The remainder of this paper is organized as follows: Section~\ref{sec: Literature Review} reviews existing fall detection systems. Section~\ref{Sec: Methodology} outlines the methodology and model development process. Section~\ref{sec:Analysis} presents experimental analyses, and Section~\ref{sec: Conclusion} concludes with directions for future research.

\section{Literature Review}\label{sec: Literature Review}

Existing research underscores the need for privacy-preserving, vision-based FDS capable of automatically alerting caregivers or emergency services. Among various approaches, thermal sensors stand out for their privacy-preserving features, robustness in diverse lighting conditions, and ability to focus on heat-emitting objects while ignoring irrelevant background details~\cite{mehta2021motion, alam2022vision}. However, effective FDS using thermal imaging must address challenges such as limited datasets, real-time processing requirements, and false alarm reduction to achieve widespread adoption. The most widely used dataset, TSF~\cite{mehta2021motion}, suffers from minimal actor diversity, constrained environments, and small sample sizes, limiting the development of generalizable solutions. Table~\ref{tab:lit-unsupervised} summarizes notable works on TSF. \cite{nogas2018fall} introduced a Convolutional LSTM Autoencoder (ConvLSTM-AE) that combined spatial and temporal features, achieving an AUC of $83\%$. This was improved by a 3D Convolutional Autoencoder (3DCAE)~\cite{nogas2020deepfall}, which leveraged contiguous frame windows for anomaly scoring, achieving $97\%$ ROC-AUC. Similarly, \cite{elshwemy2020new} achieved $97\%$ ROC-AUC with a Spatiotemporal Residual Autoencoder (SRAE) incorporating ConvLSTM layers and residual connections. \cite{khan2021spatio} explored spatiotemporal adversarial networks, achieving $95\%$ ROC-AUC with increased computational complexity. \cite{mehta2021motion} combined thermal data with motion flow through a dual-channel Autoencoder, yielding $93\%$ ROC-AUC despite challenges with region-of-interest extraction. \cite{silver2023novel} introduced a supervised approach combining a 3D CNN with an Autoencoder, outperforming individual models but yielding mediocre results due to suboptimal sub-model optimization.

These studies highlight both progress and persistent challenges in thermal fall detection, where most approaches treat fall detection as an anomaly detection problem, relying on sparse datasets with limited real-world applicability. 
This work addresses these shortcomings by pragmatically developing multiple models and conducting extensive ablation studies on the newly emerged TF-66~\cite{silver2025thermal} dataset, which contains diverse samples gathered from multiple environments and actors with varied demographics, to identify the top-performing architectures.
These selected models are then evaluated on the TSF~\cite{mehta2021motion} dataset to benchmark their performance against existing state-of-the-art methods, ensuring both strong generalization and compatibility with prior research.


\section{Methodology}\label{Sec: Methodology}

\begin{figure*}[!ht]
\centering
  \includegraphics[trim={0cm, 3.2cm, 0cm, 0cm}, clip, width=0.9\textwidth]{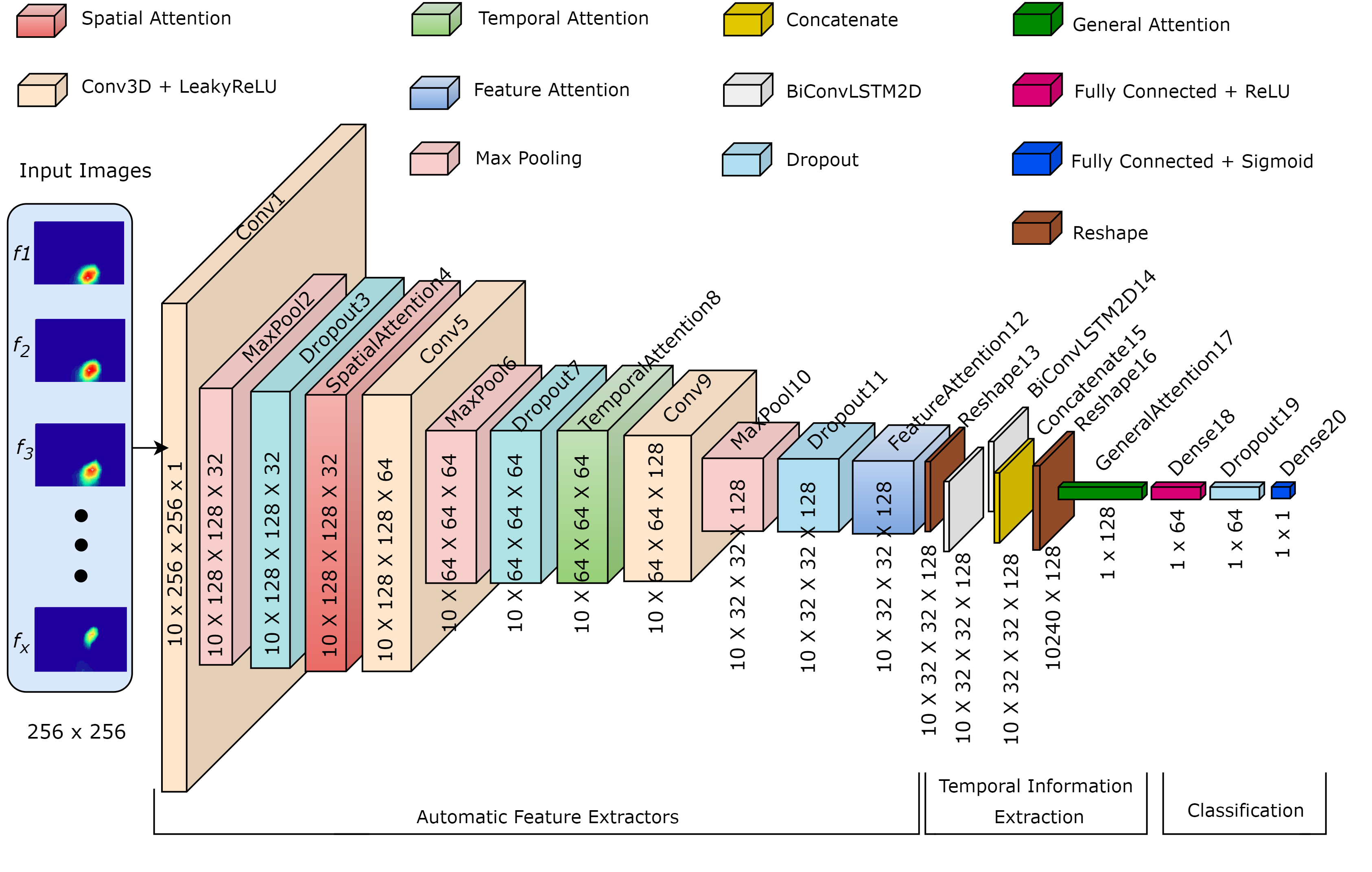}
  \caption{An illustration of the proposed attention-enhanced 3D convolutional recurrent architecture. This model extends a vanilla 3D CNN (cf.~\cref{tab:Arch:vanilla}) by integrating attention modules, including a spatial attention layer, temporal attention layer, and feature attention layer, and a BiConvLSTM2D module. 
  }
  \label{fig:BiConvLSTMAtt}
  \vskip -0.2in
\end{figure*}

We began with a vanilla 3D-CNN model, inspired by \cite{silver2023novel} and summarized in Table~\ref{tab:Arch:vanilla}, which served as the baseline.
Using a bottom-up approach, the baseline was progressively refined through the integration of sophisticated components, including spatial, temporal, and feature-based attention mechanisms, as well as self- and general-attention modules, optical (motion) flow inputs, and recurrent layers.
Each component was evaluated independently and in various combinations through controlled experiments, leading to the identification of four top-performing architectures  detailed in subsequent sections.
Initial tests indicated that a shallower 3D-CNN, with its depth illustrated in Fig.~\ref{fig:BiConvLSTMAtt}, was effective for the relatively small TF-66 dataset. The model utilized a $3 \times 3 \times 3$ kernel size with ``\texttt{same}'' padding to preserve spatial-temporal features. \texttt{LeakyReLU} activation and dropout rates of $0.25$ for \texttt{Conv} layers and $0.5$ for fully connected (FC) layers minimized overfitting. The \texttt{ADAM} optimizer (learning rate $0.0001$) and a batch size of 16 ensured stable training and consistent convergence. 


\begin{table}[!t]
\centering
\footnotesize
\caption{Architectural details of the baseline 3D-CNN}
\label{tab:Arch:vanilla}
\begin{tabularx}{\linewidth}{@{}>{\arraybackslash}p{0.25\linewidth}>{\centering\arraybackslash}p{0.25\linewidth}>{\centering\arraybackslash}p{0.40\linewidth}@{}}
\hline
\centering Layer ID & Layer Type & Output Dimension \\
\hline
Input & Input Layer & ($b$, $t$, 256, 256, ~~1) \\
Conv1+LeakyReLU & Conv3D & ($b$, $t$, 256, 256, ~32) \\
\quad MaxPool2 & MaxPooling3D & ($b$, $t$, 128, 128, ~32) \\
\quad Dropout3 & Dropout & ($b$, $t$, 128, 128, ~32) \\
Conv4+LeakyReLU &  Conv3D & ($b$, $t$, 128, 128, ~64) \\
\quad MaxPool5 & MaxPooling3D & ($b$, $t$, ~64, ~64, ~64) \\
\quad Dropout6 & Dropout & ($b$, $t$, ~64, ~64, ~64) \\
Conv7+LeakyReLU & Conv3D & ($b$, $t$, ~64, ~64, 128) \\
\quad MaxPool8 & MaxPooling3D & ($b$, $t$, ~32, ~32, 128) \\
\quad Dropout9 & Dropout & ($b$, $t$, ~32, ~32, 128) \\
Reshape10 & Reshape & ($b$, $t$, 1310720) \\
Dense11 & Dense & ($b$, 64) \\
Dropout12 & Dropout & ($b$, 64) \\
Dense13 (Output) & Dense & ($b$, 1) \\
\hline
\end{tabularx}
\begin{tabularx}{\linewidth}{@{}>{\raggedright\arraybackslash}X@{}}
Total \# of trainable parameters:  1,172,422; Memory size: 4.47 MB;\\
Activation: \{L1, L4, L7\}$\rightarrow$LeakyReLU (alpha = 0.1), Output$\rightarrow$Sigmoid; \\ Loss Function: Binary cross-entropy; Temporal Sequence Length ($t$): 10; \\
Kernel size: (3,3,3)$\rightarrow$Conv3D,   (1,2,2)$\rightarrow$MaxPooling3D layers;\\ 
Padding: always set to ``Same''; Optimizer: Adam;
Batch size ($b$): 16; \\
Dropout rate: \{L3, L6, L9\}$\rightarrow$0.25, L12$\rightarrow$0.5; Learning rate: 0.001;\\ 
\hline
\end{tabularx}
\vspace{-0.2cm}
\end{table}

\subsection{Attention Mechanisms}\label{sec-attn}

To enhance the model's ability to focus on spatial, temporal, and channel-wise features, attention mechanisms were integrated into the vanilla model, inspired by \cite{lu2018deep, achtibat2024attnlrp, su2024novel}. Following subsections introduce the basics of each attention mechanism, with further details available in \cite{woo2018cbam} for spatial attention, \cite{vaswani2017attention} for self and temporal attention, \cite{hu2018squeeze} for feature-based attention, and \cite{lu2018deep} for general attention.

\subsubsection{Spatial Attention}
The spatial attention mechanism, $\mathcal{L}_{\text{attn}}^{\text{spatial}}$, highlights relevant spatial regions in individual frames. It operates on the input feature map $\mathbf{X} \in \mathbb{R}^{B \times T \times H \times W \times C}$, where $B$ is the batch size, $T$ is the sequence length, $H$ and $W$ are frame height and width, and $C$ is the number of channels. The input is averaged along the temporal ($T$) and channel ($C$) dimensions to produce a 2D feature map $\mathbf{F}_{\text{spatial}}$, scaled by a sigmoid activation $\sigma$ and applied element-wise ($\odot$) to $\mathbf{X}$:
\begin{equation} 
\mathcal{L}_{\text{attn}}^{\text{spatial}}(\mathbf{X}) = \sigma\left(\mu_{\text{t, c}}(\mathbf{X})\right) \odot \mathbf{X}, \label{eq-spatial-attn}
\end{equation}
where $\mu_{\text{t, c}}$ averages along temporal and channel dimensions. 

\subsubsection{Temporal Attention}
Temporal attention, $\mathcal{L}_{\text{attn}}^{\text{temporal}}$, emphasizes motion dynamics (i.e., temporal regions relevant to motion) across sequences. It reduces $\mathbf{X}$ along spatial dimensions ($H$, $W$) using $\mu_{\text{h, w}}$, generating $\mathbf{F}_{\text{temporal}} \in \mathbb{R}^{B \times T \times C}$. Attention weights, scaled by $\sigma$, are applied element-wise:
\begin{equation}
\mathcal{L}_{\text{attn}}^{\text{temporal}}(\mathbf{X}) = \sigma\left(\mu_{\text{h, w}}(\mathbf{X})\right) \odot \mathbf{X}, \label{eq-temporal-attn}
\end{equation}
where $\mu_{\text{h, w}}$ averages along height and width dimensions. 

\subsubsection{Feature-Based Attention}
Feature-based attention, $\mathcal{L}_{\text{attn}}^{\text{f}}$, emphasizes informative channels by aggregating spatial and temporal features into a descriptor $\mathbf{z} \in \mathbb{R}^{B \times C}$. This is passed through reduction and restoration layers with weights $W_{\text{rdu}}$, $W_{\text{rto}}$ and biases $b_{\text{rdu}}$, $b_{\text{rto}}$, scaled by $\sigma$, and applied via $\odot$:
\begin{equation}
\mathcal{L}_{\text{attn}}^f(\mathbf{X}) = \mathbf{X} \odot \sigma\big(W_{\text{rto}} \cdot \varphi \big( W_{\text{rdu}} \mathbf{z} + b_{\text{rdu}} \big) + b_{\text{rto}} \big), \label{eq-feature-attn}
\end{equation}
where $\varphi$ is ReLU. A reduction ratio of 32 provided the optimal balance between computational efficiency and performance.

\subsubsection{Self-Attention}
Self-attention, $\mathcal{L}_{\text{attn}}^{\text{self}}$, captures global dependencies by reshaping $\mathbf{X}$ into $\mathbf{X}_{\text{reshaped}} \in \mathbb{R}^{B \times (T \cdot H \cdot W) \times C}$. Using multi-head attention (MHA), $\mathbf{X}$ is projected into query, key, and value matrices ($\mathbf{Q}$, $\mathbf{K}$, $\mathbf{V}$) for each attention head:
\begin{equation}
\text{head}_i = \text{Softmax}\left(\frac{\mathbf{Q}_i \mathbf{K}_i^\top}{\sqrt{d_k}}\right)\mathbf{V}_i, \label{eq-head-concise}
\end{equation}
where $d_k$ is the key dimension. Outputs are concatenated, projected, and reshaped:
\begin{equation}
\mathcal{L}_{\text{attn}}^{\text{self}}(\mathbf{X}) = \text{Reshape}(\text{Concat}(\text{head}_1, \dots, \text{head}_h) W_O). \label{eq-mha-concise}
\end{equation}
In this work, ablation studies identified 2 heads and a key dimension of 128 as optimal.

\subsubsection{General Attention}
General attention, $\mathcal{L}_{\text{attn}}^{\text{general}}$, refines temporal features using weights $W$ and $b$, scaled by $\text{tanh}$ ($\tau$) and softmax ($\phi$):
\begin{equation}
\mathcal{L}_{\text{attn}}^{\text{general}}(\mathbf{X}) = \sum_{t, h, w} \left(\phi\left(\tau\left(\mathbf{X}W + b\right)\right) \odot \mathbf{X}\right), \label{eq-general-attn}
\end{equation}
It ensures computational efficiency and interpretability.
Inspired by \cite{bertasius2021space}, the spatial, temporal, and feature/self-attention were applied after the first, second, and third \texttt{Conv3D} blocks, respectively, enhancing the model’s focus on spatial, temporal, and channel features. Note that \texttt{Conv3D} block refers to the following layers connected sequentially \texttt{Conv3D + LeakyReLU, MaxPool}, and \texttt{Dropout}.


\subsection{Temporal Feature Learning}\label{sec-temp-attn}

Recurrent architectures, such as \texttt{LSTM}, \texttt{Bi-LSTM}, and \texttt{ConvLSTM}, are effective for capturing spatiotemporal features in video sequences, with \texttt{ConvLSTM} excelling in this domain~\cite{su2024novel, lu2018deep}. 

The recurrent modules were placed after the final attention layer and before the \texttt{FC} layers to extract fine-grained temporal details. Attention mechanisms spanning the input sequence were also tested in combination with recurrent modules to evaluate their impact on temporal feature learning and compatibility with the baseline model. To achieve bidirectional processing, the \texttt{BiConvLSTM} layer combines forward and backward \texttt{ConvLSTM2D} layers through concatenation, enabling the model to capture temporal dependencies in both directions from input sequences. 

\subsection{Incorporating External Motion Features}\label{sec-opt-flow}

To improve robustness against illumination changes and environmental variations, such as room dimensions, motion-specific information was incorporated as an additional input channel alongside thermal signatures. Dense motion feature maps were generated using the Farneback method~\cite{farneback2003two}, which is resilient to lighting changes and ignores non-heat-emitting objects. To reduce computational overhead, motion maps were extracted offline. This approach complemented static frame information, enhancing the model’s ability to capture motion dynamics and improving fall detection performance~\cite{mehta2021motion}.

\subsection{The Integration}\label{sec-integration}

To identify the best model, advanced techniques, including recurrent modules (\texttt{ConvLSTM}, \texttt{Bi-LSTM}), attention mechanisms (spatial, temporal, feature, and self-attention), and motion features, were systematically integrated with the baseline (Vanilla 3D-CNN. Cf.~Table~\ref{tab:Arch:vanilla}). During the ablation study, each technique was incrementally added or removed in a controlled manner, ensuring a comprehensive evaluation of all possible combinations. The four top-performing configurations on the TF-66 dataset were:

\textbf{ConvLSTM + General Attention + Motion Flow (M1):}
A \texttt{ConvLSTM} layer was added after the final \texttt{Conv3D} layer, followed by a global general attention mechanism. Motion information was integrated as an additional input channel alongside thermal imaging (cf.~\cref{sec-opt-flow}).

\textbf{BiConvLSTM + Layer-specific Attention (M2):} 
Building upon M1, \texttt{ConvLSTM} was replaced with \texttt{BiConvLSTM}, excluding motion flow. Hence, spatial, temporal, and feature attention mechanisms were applied after the first, second, and third \texttt{Conv3D} convolution blocks, respectively. 

\textbf{ConvLSTM + Layer-specific Attention (M3): } 
Similar to M2, but it replaces the \texttt{BiConvLSTM} with a standard \texttt{ConvLSTM}.

\textbf{Baseline Model + Self-Attention (M4):} 
A self-attention mechanism was added after the third \texttt{Conv3D} layer and before the \texttt{FC} layer in the baseline model.

These top model configurations were then further evaluated on the TSF~\cite{mehta2021motion} benchmark dataset. Model performances on both datasets are discussed in \cref{sec-Quant}.

\section{Experimental Setup and Analysis}\label{sec:Analysis}

\subsection{Environment}\label{sec:environment}

Model development was conducted using Python 3.10.11, leveraging open-source libraries and deep learning frameworks, specifically Keras with a TensorFlow backend. Training and evaluation were performed on the Cedar computing cluster of the Digital Research Alliance of Canada, utilizing an NVIDIA Tesla K80 GPU (2,496 CUDA cores, 12GB VRAM).

\subsection{Datasets}

\begin{table}[!tp]
\centering
\caption{Dataset Summary: TF-66 vs TSF~\cite{mehta2021motion}}
\label{tab:dataset_description}
\begin{small}
\begin{sc}
\setlength{\tabcolsep}{4pt} 
\begin{tabular}{cccccccc}
\hline
Dataset & Resolution & \fall[0.75em] & {\footnotesize \faBed} & {\footnotesize \faUsers} & {\footnotesize \faVideo} &  {\ceilingmount[0.75em]} & {\footnotesize \faLock} \\ \hline
TF-66~\cite{silver2025thermal} & 140$\times$60 & 562 & 250 & 66 & 9 & \checkmark & \checkmark\\ 
TSF~\cite{mehta2021motion} & 480$\times$640 & 35 & 9 & 1 & 1 & & \\\hline
\end{tabular}
\end{sc}
\vspace{-0.15cm}
\begin{flushleft} \footnotesize
\faUsers, \faVideo, \fall[0.75em], \faBed, \ceilingmount[0.75em], and \faLock~ denote the number of participants, recording environments, the number of fall samples, of non-fall samples, the use of ceiling-mounted sensors, and the privacy-preserving nature, respectively. 
\end{flushleft}
\end{small}
\vspace{-0.5cm}
\end{table}

\textbf{TF-66:} It is the first publicly available, occlusion-free, privacy-preserving thermal dataset for fall detection, recorded in diverse real-world environments. It contains 562 fall videos and 250 non-fall videos from 66 participants, captured using a ceiling-mounted Calumino Thermal Sensor (CTS) Evaluation Kit (EVK)\footnote{\url{https://calumino.com/evaluation-kit/}} which captures videos at a resolution of 140$\times$60 at 4 frames per second (fps) across 9 environments with 3 varying room heights. Designed as a new benchmark for thermal fall detection, TF-66 incorporates evidence-based fall distributions, ensuring realistic simulations guided by a physiotherapist. To promote fair model comparisons, a standardized data generator is provided, balancing fall and non-fall samples while mitigating dataset biases. 
The dataset also provides tailored subsets reflecting deployment conditions (e.g., height-based filtering, a senior-specific subset, and a hospital subset), and maintains a fixed thermal range to avoid dynamic rescaling issues. All experiments in this work used the predefined mutually exclusive 80:20 train/validation split released with the dataset \cite{silver2025thermal}, which balances both the number of videos and the total frame counts. Researchers are encouraged to adopt this split for consistency and fair comparison. Publicly available for non-commercial research, TF-66 enables robust and reproducible evaluation of real-world deployable fall detection models.

\textbf{TSF~\cite{mehta2021motion}:} It contains only 35 fall videos and 9 non-fall videos, recorded in a single environment with one participant using a wall-mounted FLIR camera, resulting in substantial class imbalance and limited generalizability. Since no predefined split exists, we created an 80:20 train/validation division. Videos were first labeled by action type (e.g., falls from standing, walking, or sitting, as well as non-fall activities) and then randomly assigned to ensure balanced representation across both sets. For reproducibility, the exact file lists for this split are provided in our GitHub repository. Captured at a resolution of $480\times640$ under visible light, TSF lacks the environmental diversity and thermal consistency needed for practical deployment.

Table~\ref{tab:dataset_description} summarizes the key characteristics of the TF-66 and TSF datasets used in this work.

\begin{figure*}[!ht]
  \centering
  \begin{tabular}{cc}
    \includegraphics[trim={0.2cm, 0.3cm, 0cm, 0.2cm}, clip, width=0.38\textwidth]{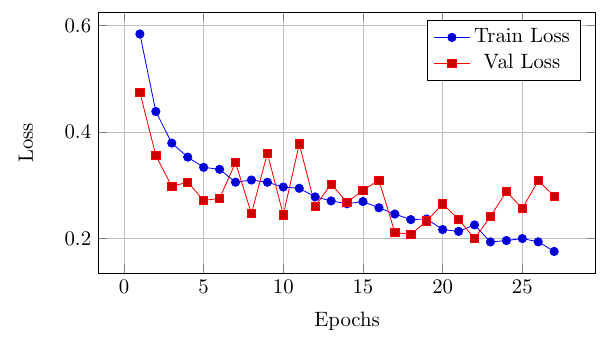} &
    \includegraphics[trim={0cm, 0.3cm, 0cm, 0.2cm}, clip, width=0.58\textwidth]{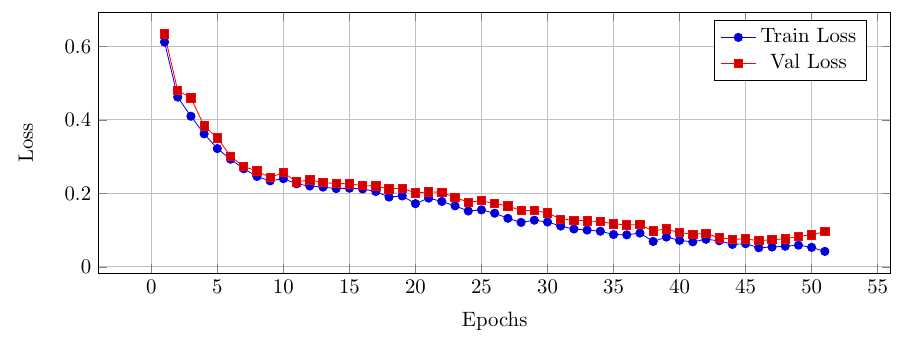} \\
    
    \includegraphics[trim={0.2cm, 0.3cmcm, 0cm, 0.2cm}, clip, width=0.38\textwidth]{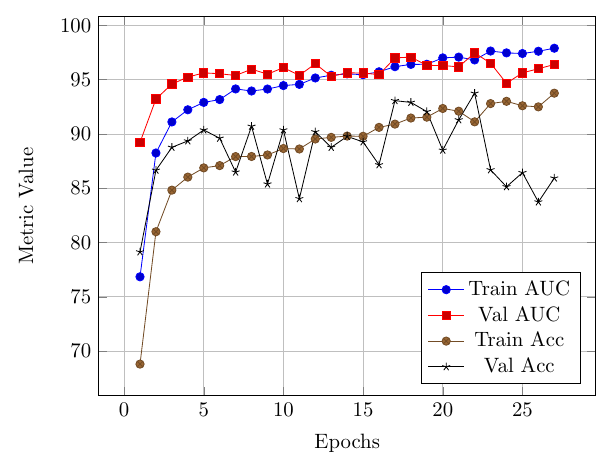} &
    \includegraphics[trim={0cm, 0.3cmcm, 0cm, 0.2cm}, clip, width=0.58\textwidth]{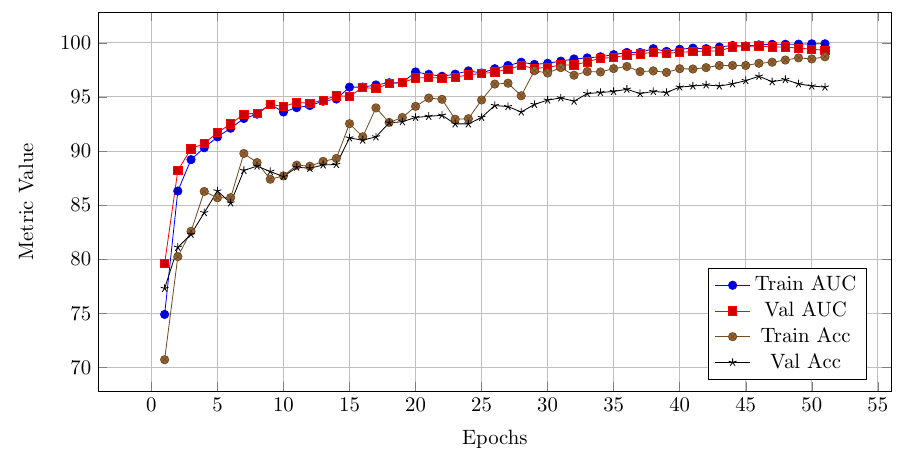} \\
    
    \includegraphics[trim={0.2cm, 0.3cm, 0cm, 0.2cm}, clip, width=0.38\textwidth]{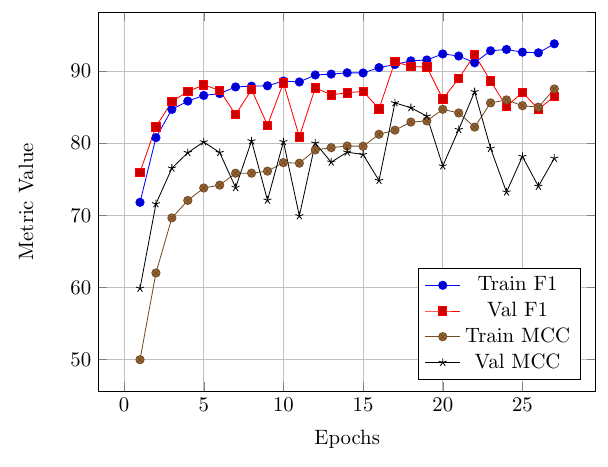} &
    \includegraphics[trim={0cm, 0.3cm, 0cm, 0.2cm}, clip, width=0.58\textwidth]{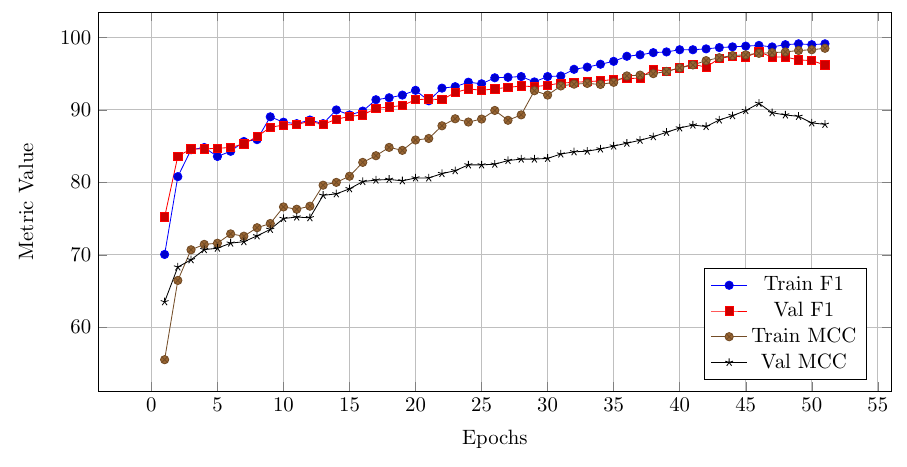} \\
  \end{tabular}
  \vspace{-0.2cm}
    \caption{Training performance of the BiConvLSTM + Layer-specific Attention (M2) model on the TF-66 (left column) and TSF (right column) datasets. The first row shows loss curves, the second row presents AUC and accuracy metrics, and the third row depicts F1 score and MCC trends.}
\label{fig:trainingplots}

\end{figure*}

\subsection{Training Strategy}

Training was conducted with early stopping applied based on validation loss, with a patience of five epochs. Each dataset (TF-66 and TSF) underwent separate end-to-end training runs, ensuring independent optimization and reproducibility. The same model architectures, hyperparameters, and data generators were used for both datasets, with only the sampling interval adjusted to account for the higher frame rate of TSF (every third frame was selected to produce same-length 10-frame sequences). The optimal models were achieved at epoch 22 and 46 for TF-66 and TSF, respectively, as shown in Fig.~\ref{fig:trainingplots}. These plots illustrate the distinct learning dynamics observed across the two datasets.

\subsection{Evaluation Metrics}\label{sec-evaluation-metrics}

Model performance was primarily evaluated using the area under the ROC curve (ROC-AUC). Additional metrics, viz. accuracy, F1-score, and Matthews Correlation Coefficient (MCC), provide complementary insights: accuracy captures overall correctness, F1-score balances precision and recall, and MCC remains reliable under class imbalance. Together, these standard metrics ensure a comprehensive evaluation. Formal definitions are omitted for brevity.

\begin{table*}[!t]
\centering
\footnotesize
\caption{Results summary of the various 3D-CNN configurations developed in this work on TF-66~\cite{silver2025thermal} and TSF~\cite{mehta2021motion}. The highlighted rows summarize the average performance metrics across both datasets for each model configuration}\label{tab:AllResults}
\setlength{\tabcolsep}{5pt} 
\begin{tabular}{m{6.5cm}lcccccccc} 
\hline
{Model} & {Dataset} & {Loss $\downarrow$}& {AUC  \%$\uparrow$} &{ACC \%$\uparrow$} & {FS \%$\uparrow$} &{MCC \%$\uparrow$}   & {GFLOPS$\downarrow$} & {PSIT} ($ms$)$\downarrow$ \\
\hline\hline
\multirow{3}{*}{Baseline (vanilla 3D-CNN)} & TF-66   & 0.381 & 93.8 & 86.7 & 84.9 & 72.6 & 37.6 & 21.0  \\
 & TSF   & 0.615 & 97.4  & 86.7 & 90.6 & 71.9 & 37.6 & 22.0 \\
 \rowcolor{green!12}
 \cellcolor{white} & Overall & 0.498 & 95.6 & 86.7 & 87.8 & 72.2 & 37.6 & 21.5\\
 \arrayrulecolor{gray}\hline
\multirow{3}{*}{ConvLSTM + General Attention + Motion Flow (M1)} & TF-66   & 0.200 & 97.5 & 93.8 & 92.3 & 87.1 & 76.3 & 243.0 \\
 & TSF     & 0.410 & 89.7 & 84.6 & 89.9 & 59.2 & 76.3 & 243.0\\
\rowcolor{green!12}
 \cellcolor{white} & Overall & 0.305 & 93.6 & 89.2 & 91.1 & 73.2 & 76.3 & 243.0\\
 \arrayrulecolor{gray}\hline
\multirow{3}{*}{BiConvLSTM + Layer-specific Attention (M2)} & TF-66   & 0.225 & 97.4 & 90.5 & 89.2 & 81.0 & 38.6 & 25.0 \\
 & TSF   & 0.071 & 99.7 & 96.9 & 98.0 & 90.9 & 38.6 & 24.0 \\
  \rowcolor{green!12}
 \cellcolor{white} & Overall & \textbf{0.148} & \textbf{97.8} & \textbf{94.3} & \textbf{94.5} & \textbf{86.2} & \underline{38.6} & \underline{24.5} \\
  \arrayrulecolor{gray}\hline
\multirow{3}{*}{ConvLSTM + Layer-specific Attention (M3)} & TF-66  &  0.205 & 96.8 & 93.2 & 90.6 & 86.3 & 38.3 & 22.0\\
 & TSF    & 0.227 & 96.9 & 90.6 & 93.5 & 76.8  & 38.3 & 23.0 \\
\rowcolor{green!12}
 \cellcolor{white} & Overall & \underline{0.226} & \underline{97.2} & \underline{90.6} & \underline{91.4} & \underline{78.9} & \textbf{38.3} & \textbf{22.5} \\
 \arrayrulecolor{gray}\hline
\multirow{3}{*}{Baseline + Self Attention (M4)} & TF-66  & 0.185 & 97.9 & 92.4 &  89.4 & 83.5 & 148.7 & 87.0 \\
 & TSF   & 0.402 &  94.0 & 82.3 & 87.4 & 61.7  & 148.7 & 86.0 \\ 
\rowcolor{green!12}
 \cellcolor{white} & Overall& 0.294 & 96.0 & 87.4 & 88.4 & 72.6 & 148.7 & 86.5 \\
 \arrayrulecolor{black}\hline\hline
\end{tabular} 
\vspace{-.1cm}
\begin{flushleft} \footnotesize
ACC – Accuracy; AUC – Area under ROC; FS – F1 Score; GFLOPS – Giga floating operations/sec; Overall – Average performance on TF-66 and TSF; PSIT – Per-sample inference time ($ms$); $\uparrow$ – higher is better; $\downarrow$ – lower is better. Boldface indicates the best result, underline indicates the second-best.  \end{flushleft}
\vspace{-.1cm}
\end{table*}

\begin{table*}[!ht]
\centering
\footnotesize
\caption{Comparative analysis of various models on the TSF Dataset~\cite{mehta2021motion}.}\label{tab:roc_results_chap5} 
\setlength{\tabcolsep}{10pt} 

\begin{tabular}{lccccc} \hline
\hspace{2cm} Model & AUC $\% \uparrow$ & \% Improvement & GFLOPS $\downarrow$ & PSIT ($ms$)$\downarrow$ & Year\\ \hline\hline
CAE Deconv.~\cite{nogas2018fall}  &   75.0   & TSF-Baseline & - & - & 2018 \\
DAE~\cite{nogas2018fall}  &  64.0   & $ - $ 14.67 & - & - & 2018  \\ 
ConvLSTM-AE ($\mu$)~\cite{nogas2018fall}  & 76.0  & $+$~~1.33 & - & - & 2018 \\
ConvLSTM-AE ($\sigma$)~\cite{nogas2018fall}  & 83.0   & $ + $ 10.67 & - & - & 2018\\ 
CLSTMAE~\cite{elshwemy2020new}  & 83.0   & $ + $ 10.67 & - & - & 2020\\ 
SRAE~\cite{elshwemy2020new} & \underline{97.0}  & \underline{$ + $ 29.33} & - & - & 2020\\ 
DSTCAE-C3D ($\mu$)~\cite{nogas2020deepfall}  & 93.0  & $ + $ 24.00 & - & - & 2020\\
DSTCAE-C3D ($\sigma$)~\cite{nogas2020deepfall}  & \underline{97.0}   & \underline{$ + $ 29.33} & - & - & 2020\\ 
Adversarial learning ($\mu$)~\cite{khan2021spatio} & 95.0  & $ + $ 26.67 & -& - & 2021\\ 
Adversarial learning ($\sigma$)~\cite{khan2021spatio} & 95.0  & $ + $ 26.67 & -& - & 2021\\
Fusion-Diff-ROI-3DCAE ($\mu$)~\cite{mehta2021motion} & 93.0  & $ + $ 24.00 & -& - & 2021\\ 
Fusion-Diff-ROI-3DCAE ($\sigma$)~\cite{mehta2021motion} & 93.0  & $ + $ 24.00 & -& - & 2021\\ 
3D CNN \cite{silver2023novel} & 79.0  & $ + $ ~~5.33 & \underline{17.70} & - & 2023\\ 
AE \cite{silver2023novel} & 74.0  & $ - $ ~~1.33 & ~~\textbf{4.03} & - & 2023 \\ 
3D CNN-AE \cite{silver2023novel} & 83.0 & $ + $ 10.67 & 21.76 & - & 2023\\ 
\rowcolor{green!15}
BiConvLSTM + Layer-specific Attention (M2 - this work) & \textbf{99.7} & \textbf{$ + $ 32.93} &  38.60 & 24 & 2025\\ \hline\hline
\end{tabular}
\vspace{-.1cm}
\begin{flushleft} \footnotesize
AUC - Area under ROC; GFLOPS - Giga floating-point operations$/$sec; PSIT - Per sample inference time; $\uparrow$ – higher is better; $\downarrow$ – lower is better. Boldface indicates the best result, underline indicates the 2nd-best; For unsupervised autoencoder models, falls are detected from reconstruction error. $\mu$ and $\sigma$ indicate whether the mean ($\mu$) or standard deviation ($\sigma$) of frame-wise errors was used as the anomaly score over a temporal window.\end{flushleft}
\vspace{-.2cm}
\end{table*}

\begin{table*}[!tp]
\centering \footnotesize
\caption{Performance of the baseline and the proposed model on the various subsets of the TF-66 dataset~\cite{silver2025thermal}}
\label{tab:SubsetResults}
\setlength{\tabcolsep}{5pt} 
\begin{tabular}{m{5.6cm}|c|cc|cc|cc|cc} 
\hline
\hspace{2cm} Model & Subset & AUC \%$\uparrow$ & \% I. & ACC \%$\uparrow$ & \% I. & FS \%$\uparrow$ & \% I. & MCC \%$\uparrow$ & \% I.\\
\hline\hline
Baseline  (vanilla 3D-CNN) & \multirow{2}{*}{Full TF-66}  & 92.9 & - & 84.5& -  & 80.4 & -  & 67.6 & -\\
BiConvLSTM + Layer-specific Attention (M2) &  & \textbf{97.4} & $ + $4.38 & \textbf{90.5} & $ + $~~7.10 & \textbf{89.2} & $ + $10.95 & \textbf{81.0} & $ + $19.82\\\hline
Baseline  (vanilla 3D-CNN)& \multirow{2}{*}{$10'$} & 95.2 & - &  90.7 & - & 90.6 & - & 82.1 & - \\
BiConvLSTM + Layer-specific Attention (M2) & & \textbf{98.4} & $ + $3.36 & \textbf{93.6} & $ + $~~3.20 & \textbf{92.1} & $ + $~~1.66 & \textbf{87.2} & $ + $~~6.21 \\\hline
Baseline  (vanilla 3D-CNN)& \multirow{2}{*}{$9'$}  & 90.2 & - & 75.6 & - & 79.7 & - & 55.1 & -\\ 
BiConvLSTM + Layer-specific Attention (M2) &  & \textbf{99.1} & $ + $9.87 & \textbf{96.7} & $ + $27.91 & \textbf{97.0} & $ + $21.71 & \textbf{93.6} & $ + $69.87\\ \hline
Baseline  (vanilla 3D-CNN)& \multirow{2}{*}{$8'$} & 89.6 & - & 81.5 & - & 73.2 & - & 59.5 & -\\
BiConvLSTM + Layer-specific Attention (M2) &  & \textbf{96.5} & $ + $7.70 & \textbf{91.2} & $ + $11.90 & \textbf{86.0} & $ + $17.49 & \textbf{79.7} & $ + $33.95\\
\hline
Baseline  (vanilla 3D-CNN)& \multirow{2}{*}{\faClinicMedical}  & \textbf{98.5} & - & \textbf{93.7} & - & \textbf{83.4} & - & \textbf{79.7} & -\\
BiConvLSTM + Layer-specific Attention (M2) & & 93.4 & $ - $5.18 & 87.2 & $ - $~~6.94 & 60.4 & $ - $27.58 & 52.8 & $ - $33.75\\\hline
Baseline  (vanilla 3D-CNN)& \multirow{2}{*}{\senioricon{}} & \textbf{99.4} & - & \textbf{97.3} & - & \textbf{91.7} & - & \textbf{90.1} & - \\
BiConvLSTM + Layer-specific Attention (M2) & & 95.2 & $ - $4.23 & 95.6 & $ - $~~1.75 & 84.0 & $ - $~~8.40 & 81.6 & $ - $~~9.43\\
\hline\hline
\end{tabular}
\vspace{-0.15cm}
\begin{flushleft} \footnotesize
ACC - Accuracy; AUC - Area under ROC; FS - F1 Score, GFLOPS - Giga floating operations$/$sec.; \% I - percentage improvement; Subset - Subgroup of data samples in TF-66, \faClinicMedical~-~Hospital; \senioricon[0.9em] - Senior; $\uparrow$ – higher is better; $\downarrow$ – lower is better. Boldface indicates the best result.\end{flushleft}
\vskip -0.2in
\end{table*}

\subsection{Quantitative Analysis}\label{sec-Quant}

Table~\ref{tab:AllResults} summarizes the experimental results for the four top-performing models and the Vanilla 3D-CNN across both datasets. The BiConvLSTM + Layer-specific Attention model (M2) achieved the highest overall performance, setting a new state-of-the-art on the TSF dataset with an AUC of $99.7\%$, accuracy and F1 scores above $95\%$, and an MCC exceeding $90\%$. On the TF-66 dataset, it remained highly competitive, significantly outperforming the baseline architecture and achieving an AUC of $97.4\%$. When averaging performance across both datasets, the BiConvLSTM + Layer-specific Attention model (M2) consistently achieved the highest scores across all metrics while marginally increasing computational complexity and adding just $3~ms$ to inference time compared to the baseline.

\captionsetup[subfloat]{labelformat=empty}
\begin{figure*}[!ht]
    \centering

    \subfloat[F. 35]{\includegraphics[width=0.11\linewidth]{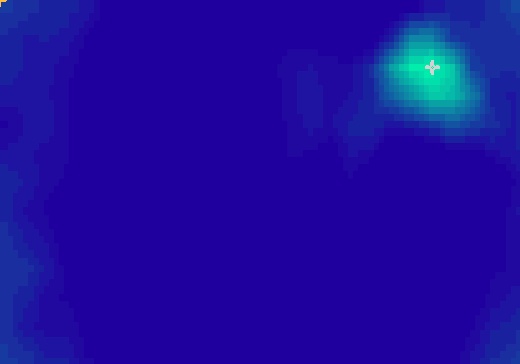}}\hfill
    \subfloat[F. 36]{\includegraphics[width=0.11\linewidth]{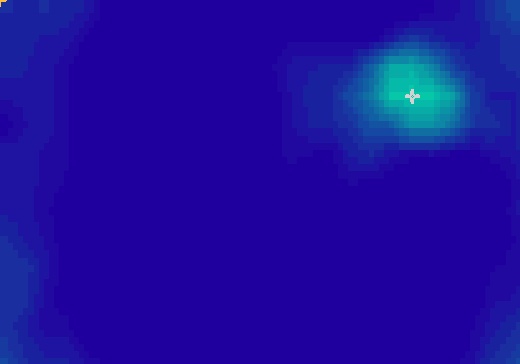}}\hfill
    \subfloat[F. 37]{\includegraphics[width=0.11\linewidth]{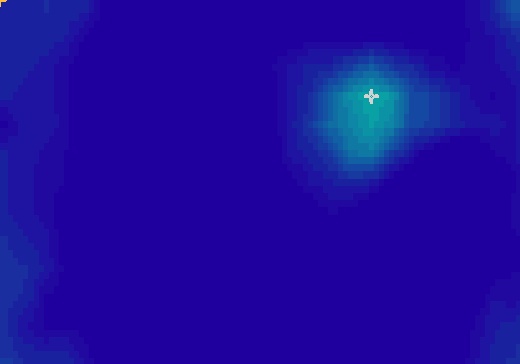}}\hfill
    \subfloat[F. 38]{\includegraphics[width=0.11\linewidth]{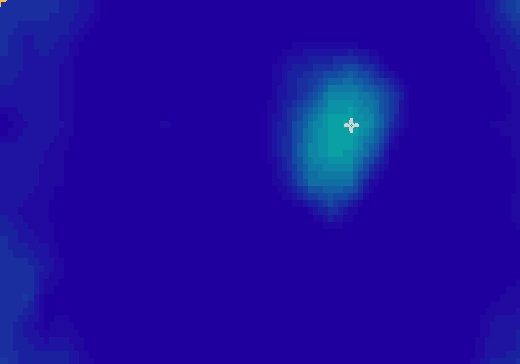}}\hfill
    \subfloat[F. 39]{\includegraphics[width=0.11\linewidth]{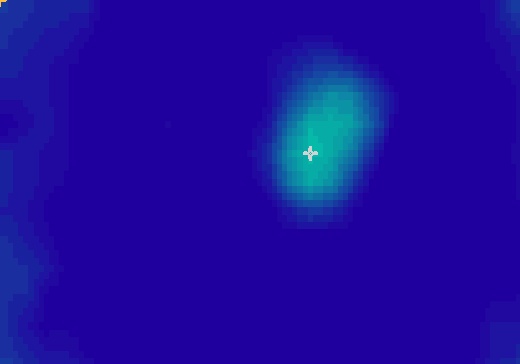}}\hfill
    \subfloat[F. 40]{\includegraphics[width=0.11\linewidth]{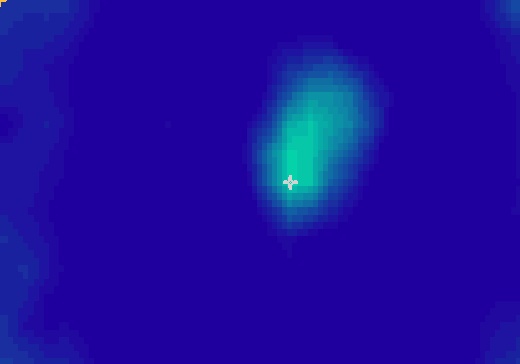}}\hfill
    \subfloat[F. 41]{\includegraphics[width=0.11\linewidth]{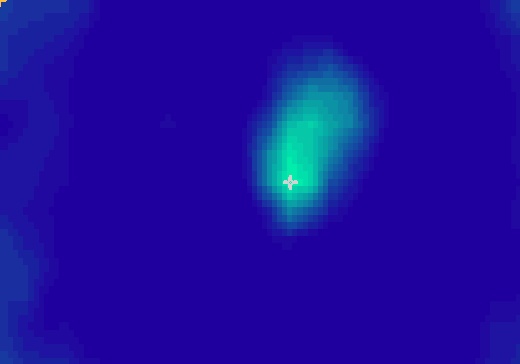}}\hfill
    \subfloat[F. 42]{\includegraphics[width=0.11\linewidth]{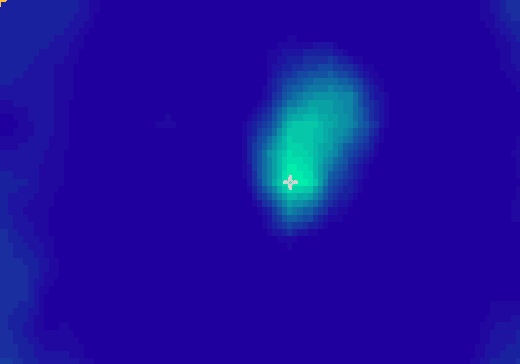}}\hfill \vspace{-.3cm}

    \subfloat[O. 35]{\includegraphics[width=0.11\linewidth]{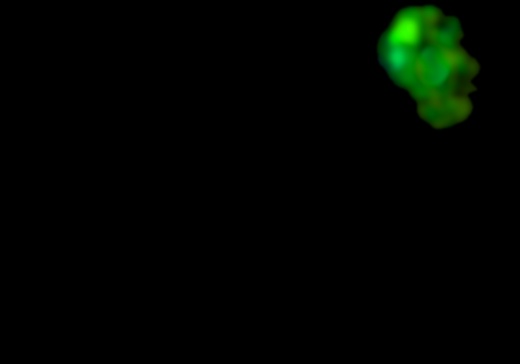}}\hfill
    \subfloat[O. 36]{\includegraphics[width=0.11\linewidth]{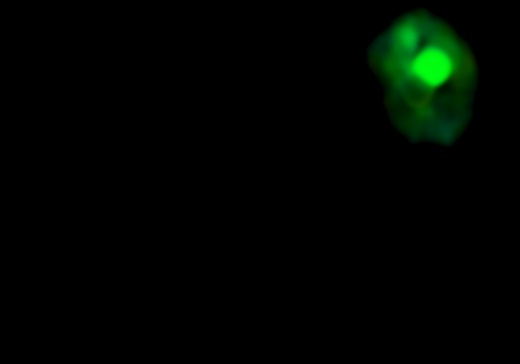}}\hfill
    \subfloat[O. 37]{\includegraphics[width=0.11\linewidth]{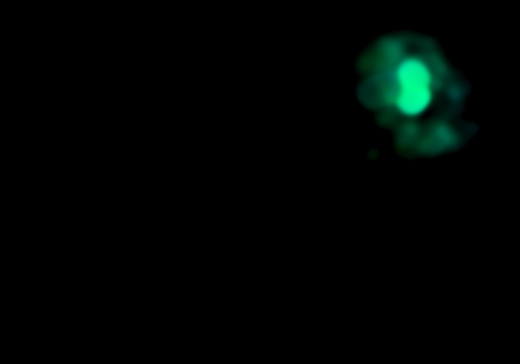}}\hfill
    \subfloat[O. 38]{\includegraphics[width=0.11\linewidth]{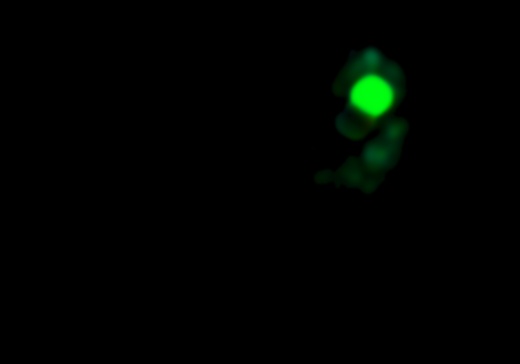}}\hfill
    \subfloat[O. 39]{\includegraphics[width=0.11\linewidth]{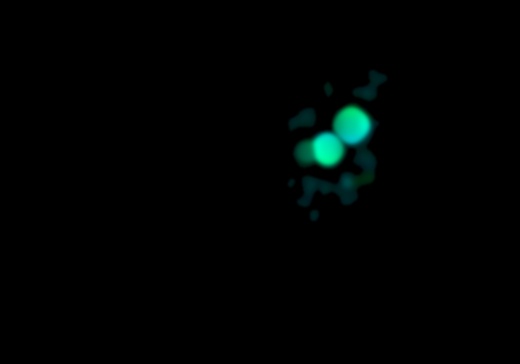}}\hfill
    \subfloat[O. 40]{\includegraphics[width=0.11\linewidth]{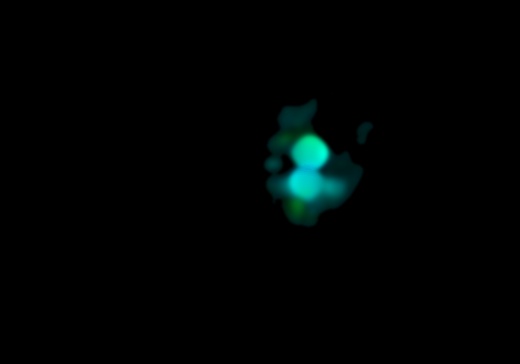}}\hfill
    \subfloat[O. 41]{\includegraphics[width=0.11\linewidth]{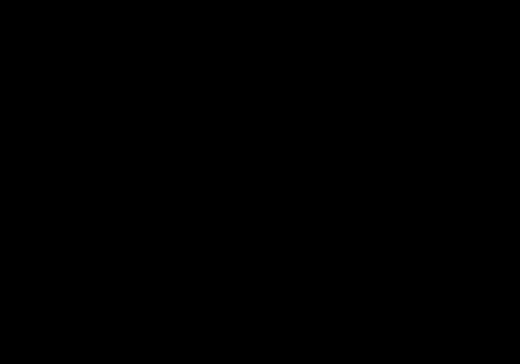}}\hfill
    \subfloat[O. 42]{\includegraphics[width=0.11\linewidth]{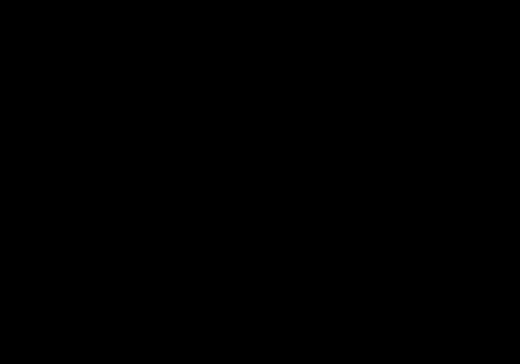}}\hfill
    \vspace{-.2cm}
    \begin{flushleft} \footnotesize
    Note: 1st row – the original images, 2nd row – the motion flow, which is the difference between the image at that time stamp and the previous time stamp. This video represents a person walking into the scene, falling onto their stomach, and remaining prone.
    \end{flushleft}
    \vspace{-0.3cm}
    \caption{Eight consecutive frames from \texttt{01-Fall-04} starting at frame 35 from the TF-66 Dataset.}
    \label{fig:correctClassification}
\end{figure*}

The ConvLSTM + General Attention + Motion Flow model (M1), optimized for TF-66, performed well on that dataset but poorly on TSF, likely due to data drifting. TF-66 contains thermal-only heat signatures, whereas TSF includes visible-light thermal images with environmental details. As a result, motion flow parameters tuned for TF-66 did not transfer well to TSF.
Although the ConvLSTM + General Attention + Motion Flow model
(M1) slightly outperformed the BiConvLSTM + Layer-specific Attention model (M2) on TF-66, it doubled computational complexity and approached the real-time limit. With a 4 fps capture rate, each 10-frame sample must be processed within $250~ms$. The motion flow model required $243~ms$ ($197~ms$ for flow generation, $46~ms$ for inference), leaving almost no buffer. Even minor delays from sensor latency or system load could exceed this threshold, triggering cascading bottlenecks and lag. In deployment, such delays risk fall events being detected minutes or hours late, which is unacceptable given the dangers of long-lie injuries. Thus, the marginal performance gain is outweighed by real-time risks.

\subsection{Qualitative Analysis}\label{sec:QualitativeAnalysis}

To visually evaluate the proposed model's performance, specific samples were analyzed for classification accuracy. Fig.~\ref{fig:correctClassification} shows 8 consecutive frames and corresponding motion flow representations from video \texttt{01-Fall-04}, starting at frame 35. This sample was consistently classified correctly across the top-performing models. Between frames 35 and 39, a dense, circular heat signature transitions into a dimmer, elongated signature, indicating the individual falling to the ground. This elongated signature persists in frames 40 to 42, representing the individual lying prone rather than bending over. The motion flow data corroborates this, with high motion flow intensities during the fall (frames 35 to 39) and minimal activity in later frames (frames 40 to 42) as the individual remains stationary. This analysis highlights the model's ability to effectively integrate spatial and temporal features from thermal imagery and motion flow, demonstrating robust and accurate fall detection.

\subsection{Complexity Analysis}\label{sec-discuss}

While all top-performing models achieved competitive ROC-AUC, those incorporating motion flow (M1) or self-attention modules (M4) had significantly higher GFLOPs than the baseline, with motion flow doubling and self-attention quadrupling computational costs. Despite this, these techniques provided substantial performance gains, outperforming hundreds of baseline-based configurations. Per-sample inference times (PSIT) ranged from $20$ to $25~ms$ for most models, except those with motion flow and self-attention, which required longer due to increased GFLOPs. Nevertheless, all models remained within the $250~ms$ real-time threshold for processing a 10-frame sample, ensuring that no rapid fall events are missed, given the system’s 4 FPS frame rate and overlapping sequence processing. However, as previously discussed, the motion flow (M1) model’s $243~ms$ inference time left little room for variability, making it an impractical choice for real-world deployment.

It is important to note that the proposed models are intended for cloud-based deployment rather than on-device edge inference. This design choice reflects the computational demands of high-performing models and the clinical requirement to minimize both missed falls and false alarms, while still achieving real-time processing at 4~fps using cloud-based deployment.

To limit overhead, all attention blocks are implemented with global pooling and pointwise gating (spatial/temporal) or squeeze–excitation with a reduction ratio of 32 (channel), and are applied after $3\times$ spatial downsampling (256$\to$32). The BiConvLSTM uses 64 filters with $3{\times}3$ kernels over $T{=}10$ frames on the reduced 32$\times$32 feature maps. These design choices explain the small increase from the baseline (37.6 GFLOPs; 21--22\,ms) to M2/M3 (38.6/38.3 GFLOPs; $\approx$22--25\,ms), while motion flow and self-attention incur much larger costs.

\subsection{Optimal Models and Real-World Implications} \label{sec-optimalModel}

The best model, BiConvLSTM + Layer-specific Attention (M2), integrates a \texttt{BiConvLSTM} layer with layer-specific attention mechanisms. Its architecture is illustrated in Fig.~\ref{fig:BiConvLSTMAtt}, where spatial, temporal, and feature attention mechanisms are applied sequentially after the first, second, and third \texttt{Conv3D} layers, respectively, followed by a \texttt{BiConvLSTM} layer and a global attention mechanism before the flatten layer. This architecture enhances feature extraction, achieving state-of-the-art performance with an ROC-AUC of $99.7\%$ on TSF as illustrated in Table \ref{tab:roc_results_chap5}, surpassing all previous models, including those reviewed in the literature. On the full TF-66 dataset, the model achieved an ROC-AUC of $97.4\%$, setting a new benchmark in thermal fall detection. As previously described, TF-66 includes data subsets that can be toggled within the data generator to isolate specific conditions. Subset-specific results in \cref{tab:SubsetResults} show that models trained on 10-, 9-, and 8-foot room height subsets outperformed the baseline across all metrics. 

However, the hospital subset underperformed, likely due to its limited sample size, where a few misclassifications significantly impacted overall results. The senior subset also exhibited a slight performance decrease. Despite these reductions, overall performance improved significantly when evaluating on the entire dataset rather than on individual subsets. These results establish the BiConvLSTM + Layer-specific Attention (M2) model as a new standard for thermal fall detection, pushing performance boundaries and providing robust benchmark results for TF-66. The combination of this model and dataset offers future researchers a strong foundation for advancing the field, as TF-66 is the most diverse and robust thermal fall detection dataset available. 
With the BiConvLSTM model demonstrating state-of-the-art performance on both TF-66 and the industry-standard TSF dataset, this work sets a new precedent for privacy-preserving, real-world-deployable fall detection systems.

\section{Conclusion}\label{sec: Conclusion}

This study presents a novel approach to real-time thermal fall detection by evaluating advanced 3D convolutional recurrent architectures with motion flow and attention mechanisms. The proposed BiConvLSTM + Layer-specific Attention (M2) model achieves state-of-the-art performance on the TSF dataset and demonstrates strong generalizability on the newly introduced TF-66 benchmark.  

The results highlight trade-offs between accuracy, computational complexity, and latency, showing that real-time feasibility can be achieved with lightweight, scalable architectures. While motion flow offers robustness, its high computational cost limits practical deployment.  

Given the limitations of TSF, we recommend prioritizing TF-66, which better represents real-world conditions. Optimizing solely on TSF risks overfitting, whereas TF-66 supports the development of more reliable, privacy-preserving solutions that improve safety and autonomy for at-risk populations.  

In addition to achieving strong accuracy, the system is envisioned as an AI-assisted tool to support caregiving and help mitigate the growing shortage of personal support workers, extending the capacity of eldercare facilities while protecting resident privacy and dignity. 

This work assumes near-constant environmental conditions, as deployment is aimed at well-monitored eldercare facilities. We acknowledge this may limit applicability in uncontrolled settings. Moreover, curated datasets cannot capture all real-world factors such as occlusion, ambient heat, or background movement. To bridge this gap, pilot testing is underway in long-term care facilities, enabling collection of authentic data to refine both the dataset and the models. Future work will also focus on reducing motion flow preprocessing overhead and improving efficiency for edge deployment.

For reproducibility, all code for the best-performing model (M2), together with scripts for generating the standardized 80:20 splits, motion flow, and dataset access links, are provided at: [link removed for review].


\begin{thebibliography}{10}

\bibitem{nogas2018fall}
J.~Nogas, S.~S. Khan, and A.~Mihailidis, ``Fall detection from thermal camera using convolutional lstm autoencoder,'' in {\em Proceedings of the 2nd workshop on aging, rehabilitation and independent assisted living, IJCAI workshop}, 2018.

\bibitem{nogas2020deepfall}
J.~Nogas, S.~S. Khan, and A.~Mihailidis, ``Deepfall: Non-invasive fall detection with deep spatio-temporal convolutional autoencoders,'' {\em Journal of Healthcare Informatics Research}, vol.~4, pp.~50--70, 2020.

\bibitem{elshwemy2020new}
F.~A. Elshwemy, R.~Elbasiony, and M.~T. Saidahmed, ``A new approach for thermal vision based fall detection using residual autoencoder.,'' {\em International Journal of Intelligent Engineering \& Systems}, vol.~13, no.~2, 2020.

\bibitem{khan2021spatio}
S.~S. Khan, J.~Nogas, and A.~Mihailidis, ``Spatio-temporal adversarial learning for detecting unseen falls,'' {\em Pattern Analysis and Applications}, vol.~24, no.~1, pp.~381--391, 2021.

\bibitem{mehta2021motion}
V.~Mehta, A.~Dhall, S.~Pal, and S.~S. Khan, ``Motion and region aware adversarial learning for fall detection with thermal imaging,'' in {\em 2020 25th international conference on pattern recognition (ICPR)}, pp.~6321--6328, IEEE, 2020.

\bibitem{silver2023novel}
C.~Silver and T.~Akilan, ``A novel approach for fall detection using thermal imaging and a stacking ensemble of autoencoder and 3d-cnn models,'' in {\em 2023 IEEE Canadian Conference on Electrical and Computer Engineering (CCECE)}, pp.~71--76, IEEE, 2023.

\bibitem{yu2022data}
X.~Yu, T.~Ma, J.~Jang, and S.~Xiong, ``Data augmentation to address various rotation errors of wearable sensors for robust pre-impact fall detection,'' {\em IEEE journal of biomedical and health informatics}, vol.~27, no.~5, pp.~2197--2207, 2022.

\bibitem{chen2024fall}
X.~Chen, J.~Yan, S.~Qin, P.~Li, S.~Ning, and Y.~Liu, ``Fall detection method based on a human electrostatic field and vmd-ecanet architecture,'' {\em IEEE Journal of Biomedical and Health Informatics}, 2024.

\bibitem{chaudhuri2017older}
S.~Chaudhuri, L.~Kneale, T.~Le, E.~Phelan, D.~Rosenberg, H.~Thompson, and G.~Demiris, ``Older adults’ perceptions of fall detection devices,'' {\em Journal of applied gerontology}, vol.~36, no.~8, pp.~915--930, 2017.

\bibitem{wang2023convolution}
P.~Wang, Q.~Li, P.~Yin, Z.~Wang, Y.~Ling, R.~Gravina, and Y.~Li, ``A convolution neural network approach for fall detection based on adaptive channel selection of uwb radar signals,'' {\em Neural Computing and Applications}, vol.~35, no.~22, pp.~15967--15980, 2023.

\bibitem{alam2022vision}
E.~Alam, A.~Sufian, P.~Dutta, and M.~Leo, ``Vision-based human fall detection systems using deep learning: A review,'' {\em Computers in biology and medicine}, vol.~146, p.~105626, 2022.

\bibitem{naser2022privacy}
A.~Naser, A.~Lotfi, M.~D. Mwanje, and J.~Zhong, ``Privacy-preserving, thermal vision with human in the loop fall detection alert system,'' {\em IEEE Transactions on Human-Machine Systems}, vol.~53, pp.~164--175, 2022.

\bibitem{rafferty2019thermal}
J.~Rafferty, J.~Medina-Quero, S.~Quinn, C.~Saunders, I.~Ekerete, C.~Nugent, J.~Synnott, and M.~Garcia-Constantino, ``Thermal vision based fall detection via logical and data driven processes,'' in {\em 2019 IEEE International Conference on Big Data, Cloud Computing, Data Science \& Engineering (BCD)}, pp.~35--40, IEEE, 2019.

\bibitem{riquelme2019ehomeseniors}
F.~Riquelme, C.~Espinoza, T.~Rodenas, J.-G. Minonzio, and C.~Taramasco, ``ehomeseniors dataset: An infrared thermal sensor dataset for automatic fall detection research,'' {\em Sensors}, vol.~19, p.~4565, 2019.

\bibitem{newaz2023methods}
N.~T. Newaz and E.~Hanada, ``The methods of fall detection: A literature review,'' {\em Sensors}, vol.~23, no.~11, p.~5212, 2023.

\bibitem{pentyala2021privacy}
S.~Pentyala, R.~Dowsley, and M.~De~Cock, ``Privacy-preserving video classification with convolutional neural networks,'' in {\em ICML}, pp.~8487--8499, PMLR, 2021.

\bibitem{silver2025thermal}
C.~Silver and T.~Akilan, ``Thermal fall 66: A robust dataset for thermal imaging-based fall detection and eldercare,'' {\em Engineering Applications of Artificial Intelligence}, vol.~160, p.~111819, 2025.

\bibitem{lu2018deep}
N.~Lu, Y.~Wu, L.~Feng, and J.~Song, ``Deep learning for fall detection: Three-dimensional cnn combined with lstm on video kinematic data,'' {\em IEEE journal of biomedical and health informatics}, vol.~23, no.~1, pp.~314--323, 2018.

\bibitem{achtibat2024attnlrp}
R.~Achtibat, S.~M.~V. Hatefi, M.~Dreyer, A.~Jain, T.~Wiegand, S.~Lapuschkin, and W.~Samek, ``Attn{LRP}: Attention-aware layer-wise relevance propagation for transformers,'' in {\em Forty-first International Conference on Machine Learning}, 2024.

\bibitem{su2024novel}
C.~Su, J.~Wei, D.~Lin, L.~Kong, and Y.~L. Guan, ``A novel model for fall detection and action recognition combined lightweight 3d-cnn and convolutional lstm networks,'' {\em Pattern Analysis and Applications}, vol.~27, no.~1, p.~3, 2024.

\bibitem{woo2018cbam}
S.~Woo, J.~Park, J.-Y. Lee, and I.~S. Kweon, ``Cbam: Convolutional block attention module,'' in {\em Proceedings of the European conference on computer vision (ECCV)}, pp.~3--19, 2018.

\bibitem{vaswani2017attention}
A.~Vaswani, ``Attention is all you need,'' {\em Advances in Neural Information Processing Systems}, 2017.

\bibitem{hu2018squeeze}
J.~Hu, L.~Shen, and G.~Sun, ``Squeeze-and-excitation networks,'' in {\em Proceedings of the IEEE conference on computer vision and pattern recognition}, pp.~7132--7141, 2018.

\bibitem{bertasius2021space}
G.~Bertasius, H.~Wang, and L.~Torresani, ``Is space-time attention all you need for video understanding?,'' in {\em ICML}, vol.~2, p.~4, 2021.

\bibitem{farneback2003two}
G.~Farneb{\"a}ck, ``Two-frame motion estimation based on polynomial expansion,'' in {\em Image Analysis: 13th Scandinavian Conference, SCIA 2003 Halmstad, Sweden, June 29--July 2, 2003 Proceedings 13}, pp.~363--370, Springer, 2003.

\end{thebibliography}
\end{document}